\documentclass[letterpaper, 10 pt, conference]{ieeeconf}  %

\IEEEoverridecommandlockouts                              %

\usepackage[T1]{fontenc}                                                          

\usepackage{tikz} %

\usepackage{amsfonts}	%
\usepackage{amsmath}	%
\usepackage{amssymb}    %
\usepackage{siunitx}
\usepackage{pifont}   %
\usepackage{xspace}

\usepackage{booktabs}
\usepackage{makecell}  %
\usepackage[flushleft]{threeparttable}  %
\usepackage{multirow}
\usepackage{rotating}

\usepackage{xspace}    %
\usepackage{xcolor}    %
\let\labelindent\relax
\usepackage[inline]{enumitem} %
\usepackage{textcomp}  %
\usepackage{gensymb}   %
\usepackage{graphicx} %
\graphicspath{{figures/}}
\usepackage{microtype}
\usepackage{cite}
\usepackage{flushend}
\usepackage{cuted}
\usepackage{capt-of}  %

\usepackage{pgfplots}
\pgfplotsset{width=7cm, compat=1.10}
\usepgfplotslibrary{fillbetween}
\usepackage{tabularx}

\newcolumntype{P}[1]{>{\centering\arraybackslash}p{#1}}

\makeatletter
\let\NAT@parse\undefined
\makeatother

\usepackage{url}
\usepackage[pdfencoding=auto, breaklinks,colorlinks,hypertexnames=false]{hyperref} %
\usepackage[hang,flushmargin]{footmisc}

\usepackage[capitalize]{cleveref}
\crefname{section}{Sec.}{Secs.}
\Crefname{section}{Section}{Sections}
\Crefname{table}{Table}{Tables}
\crefname{table}{Tab.}{Tabs.}

\usepackage[nogroupskip,acronyms,nopostdot,style=super,nonumberlist,toc]{glossaries}

\usepackage{colortbl}

\newcommand{\ours}{GEMS\xspace}
\newcommand{\oursdata}{GEMS-Bench\xspace}

\newif\ifanonymous
\anonymousfalse

\usepackage{amsmath}
\usepackage{amssymb}

\DeclareMathAlphabet{\mathcal}{OMS}{cmsy}{m}{n}

\usepackage{xspace}

\renewcommand{\[}{\begin{equation}}
\renewcommand{\]}{\end{equation}}

\renewcommand{\baselinestretch}{0.99}

\definecolor{red}{RGB}{255, 0, 0}   %
\definecolor{orange}{RGB}{255, 77, 0}   %
\definecolor{green}{RGB}{0, 128, 0}   %
\definecolor{purple}{RGB}{160, 32, 240}   %
\definecolor{lightblue}{RGB}{52, 155, 235}   %
\definecolor{darkmagenta}{RGB}{204, 51, 139} %

\usepackage[capitalize]{cleveref}

\crefname{figure}{Fig.}{Figs.}
\Crefname{figure}{Figure}{Figures}
\crefname{section}{Sec.}{Secs.}
\Crefname{section}{Section}{Sections}
\Crefname{table}{Table}{Tables}
\crefname{table}{Tab.}{Tabs.}
\crefname{algorithm}{Algo.}{Algos.}
\Crefname{algorithm}{Algorithm}{Algorithms}
\crefname{appendix}{Sec.}{Secs.}
\Crefname{appendix}{Section}{Sections}

\title{\LARGE \bf
Gaussian Episodic Memory for Long-Term Spatiotemporal Reasoning
}

\ifanonymous
\author{Anonymous submission}
\else
\author{
Ling-Chia Chen\textsuperscript{*},
Abdelrhman Werby\textsuperscript{*},
Kaan Aytekin,
Kai O. Arras%
\thanks{\textsuperscript{*}These authors contributed equally to this work.}%
\thanks{L. Chen, A. Werby, K. Aytekin, and K. O. Arras are
with the Socially Intelligent Robotics Lab, Institute for
Artificial Intelligence, University of Stuttgart, Germany.
Email: {\tt\small abdelrhman.werby@ki.uni-stuttgart.de}}
\thanks{The authors thank the International Max Planck Research
School for Intelligent Systems (IMPRS-IS) for supporting
Abdelrhman Werby and Kaan Aytekin.}
}
\fi

\IEEEaftertitletext{\vspace{-1\baselineskip}\ifanonymous\else\vspace{-35pt}\fi}
\begin{document}
\bstctlcite{IEEEexample:BSTcontrol}

\maketitle
\thispagestyle{empty}
\pagestyle{empty}

\begin{strip}
    \centering
    \includegraphics[width=\textwidth]{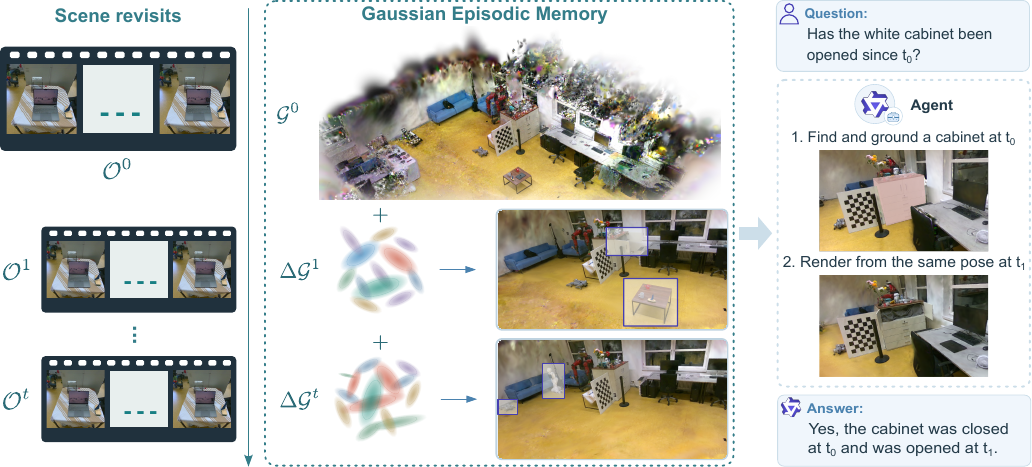}
    \captionof{figure}{\textbf{Overview of \ours.}
    A robot revisits a semi-static scene, yielding a posed RGB-D sequence $\mathcal{O}^t$ per visit (left).
    \ours reconstructs the first visit into a Gaussian map $\mathcal{G}^0$ and records each later visit as a
    difference $\Delta\mathcal{G}^t$ capturing only the regions that changed, so the scene at any past visit is
    recovered by rendering $\mathcal{G}^0$ together with the differences up to that visit (middle).
    Given a question (right), the agent explores the scene using its toolbox and answers based on the collected evidence.}
    \label{fig:teaser}
\end{strip}

\begin{abstract}
    Real-world environments undergo continual semi-static changes, requiring a memory representation that captures both the spatial and temporal evolution of a scene.
Existing spatio-temporal memories for robots typically rely either on structured representations, which are limited to the entities and attributes captured during construction, or on image-based memories, which are constrained to previously observed viewpoints.
We introduce \ours, a framework that combines a persistent spatio-temporal 3D Gaussian Splatting (3DGS) map with an agentic reasoning pipeline.
The 3DGS map is reconstructed once and subsequently updated only in regions where the scene changes, storing only these modifications while preserving visual detail across time.
The agentic pipeline leverages this map to render novel views on demand, iteratively accumulating evidence to answer spatio-temporal questions and ground objects from natural-language queries.
To advance future research, we introduce \oursdata, a spatiotemporal question-answering benchmark focused on semi-static environments. It comprises 3 real-world and 12 synthetic environments, spanning 473 question-answer pairs.
We evaluate \ours on novel-view synthesis quality across scene updates, 3D object grounding, spatiotemporal question answering, and storage efficiency. \ours outperforms recent baselines in 3D object grounding and spatiotemporal question answering, while achieving rendering quality on par with dedicated scene-update methods.
\end{abstract}
\glsresetall

\section{Introduction}
\label{sec:introduction}

Robots operating for long periods in human-centered environments must model not only the geometry and semantics of their surroundings but also how those surroundings evolve over time. Real-world environments, such as homes and offices, experience frequent semi-static changes in which objects are moved, furniture is rearranged, and equipment may appear, disappear, or change state. Modeling these changes and building episodic memory that captures when and where they occurred enables robots to reason about the environment's history and answer spatial, semantic, and temporal questions, such as where an object was previously located, when a change occurred, or how the current state of a scene differs from an earlier one. 

3D scene graphs (3DSGs)~\cite{rotondi2026survey,werby2024hierarchical,werby2025keysg} have emerged as compact, structured representations that organize spatial-semantic concepts. However, typical 3DSGs represent a single snapshot of the environment. More recent 3DSG approaches incorporate temporal information into scene representations to model semi-static changes~\cite{schmid2024khronos,behrens2025lost,saxena2025grapheqa,gorlo2026describe}. However, they either preserve only object-level states and textual descriptions, precommitting to objects segmented at construction time, or preserve observations only from fixed viewpoints, limiting access to the underlying visual evidence. 
Recently, 3D Gaussian Splatting (3DGS)~\cite{kerbl20233d} has emerged as a high-fidelity representation that preserves fine-grained visual details, supports novel-view synthesis, and enables localized updates~\cite{galappaththige2026changes,yugay2026gaussian}, making it well suited for long-term embodied reasoning. Existing 3DGS-based mapping and reasoning methods, however, either address static scenes~\cite{lu2026gsmem,yu2026splatreasoner} or maintain only the latest scene state~\cite{fu2025gs}.

To this end, we present \ours, a persistent spatiotemporal episodic memory based on 3D Gaussian Splatting (3DGS), coupled with an agentic reasoning pipeline. \ours construct an initial 3DGS map, then detect changes from observations collected during each subsequent scene revisit and locally update the Gaussian primitives in affected regions. Unchanged Gaussians are shared across temporal states, while additions, removals, and modifications are stored as temporal deltas. As a result, storage grows with environmental changes rather than the number of revisits.
Given a query, the agentic pipeline can select informative viewpoints for reconstructed regions rather than being restricted to a fixed set of viewpoints. The 3DGS-based memory enables the agent to render the map from a consistent virtual camera pose across different timesteps. This controlled alignment supports cross-temporal visual comparison and reasoning about object motion and state changes. Moreover, photorealistic novel-view rendering enables the agent to perform on-demand segmentation of target objects from informative, query-dependent viewpoints.

To foster future research, we introduce \oursdata, a multi-revisit spatiotemporal embodied question-answering benchmark. It comprises 12 synthetic household scenes and 3 real-world scenes, each with at least 3 revisits. \oursdata spans 473 question-answer pairs categorized into 5 types, as well as posed RGB-D for each revisit. 

In summary, we make the following contributions:
\begin{itemize}
    \item We introduce \ours, a 3DGS-based episodic memory framework that maintains long-term scene history via efficient, localized updates, enabling high-fidelity novel-view rendering over time.
    \item We develop an agentic EQA pipeline that utilizes spatiotemporal novel-view synthesis for query-dependent viewpoint selection, visual alignment across time, and on-demand object grounding.
    \item We introduce a new multi-visit spatiotemporal question-answering benchmark, \oursdata, comprising semi-static environmental changes across 15 scenes and 473 question-answer pairs.
    \item We evaluate \ours across four dimensions: novel-view synthesis quality under scene updates, 3D object grounding from natural-language queries, spatiotemporal question answering, and storage requirements.  
\end{itemize}
\ifanonymous
Code and benchmark will be released upon acceptance.
\else
We provide code and data at: \url{https://abwerby.github.io/GEMS/}
\fi

\section{Related Work}
\label{sec:relatedwork}

\subsection{Spatial-Semantic and Spatiotemporal Mapping}

Early works on robotics mapping~\cite{mur2017orb} focused only on geometric modeling of the environment. Subsequent works~\cite{mccormac2017semanticfusion, huang2023visual} augmented such 3D geometric maps with semantic and open-vocabulary language features, enabling object retrieval and language-guided navigation. Structured representations, such as 3D Scene Graphs (3DSGs)~\cite{armeni20193d, hughes2022hydra, gu2024conceptgraphs, werby2024hierarchical, werby2025keysg, linok2025beyondbarequeries}, have emerged as more efficient representations that encode both geometric and semantic information in a topological graph; however, they only maintain a single snapshot of the scene state.

To extend robotics maps to also capture the temporal aspect of the environment, Khronos~\cite{schmid2024khronos} performs spatio-temporal metric-semantic SLAM that jointly captures short-term dynamics and long-term changes and can be queried for the estimated scene state at any past timestamp. However, its metric-semantic map does not preserve photorealistic appearance, so visual evidence beyond that abstraction cannot be re-examined. Other methods maintain a persistent object- or point-level representation and revise it as the scene changes~\cite{yan2025dynamic, liu2025dynamem}, whereas~\cite{behrens2025lost, gorlo2026describe} additionally retain temporal traces in the form of object trajectories and timestamped descriptions, respectively. The former keep only the current state and therefore cannot answer questions about earlier configurations, while all of them precommit to the objects and attributes instantiated at construction time and discard the underlying visual details. 

\subsection{3DGS-based Maps}

3D Gaussian Splatting (3DGS)~\cite{kerbl20233d} is an explicit representation for 3D scene reconstruction that enables photorealistic rendering while preserving fine-grained visual details. Because it stores explicit primitives, 3DGS also supports cheap local updates, unlike implicit representations such as NeRF~\cite{mildenhall2020nerf}. Recent extensions~\cite{yu2024language} further enrich the Gaussian primitives with high-dimensional semantic attributes for semantic tasks.
For long-term scene updates, O-SCD~\cite{galappaththige2026changes} localizes changes by directly comparing new observations against views rendered from the previous map, and others~\cite{ackermann2025cl, yugay2026gaussian} further show the gain in time and storage from updating only the changed area.
A third line of work uses 3DGS as the map representation for embodied reasoning. SplatReasoner~\cite{yu2026splatreasoner} synthesizes query-conditioned viewpoints that reveal evidence missing from fixed-view memories, and GSMem~\cite{lu2026gsmem} renders informative viewpoints on demand for exploration; both reason over a single static capture. GS-LTS~\cite{fu2025gs} instead keeps the map current across revisits, yet stores only the latest state and discards the scene's history. \ours renders query-conditioned views over a memory spanning visits, so the agent can hold a viewpoint fixed while varying the time.

\subsection{Embodied Question Answering and Episodic Memory}

Introduced by OpenEQA~\cite{majumdar2024openeqa}, Episodic Memory EQA (EM-EQA) evaluates an agent's capability to answer questions based on past multimodal experiences. In this task, a key requirement for the agent is to correctly retrieve historical contexts during querying. ReMEmbR~\cite{anwar2025remembr} retrieves by embedding similarity over textual descriptions of video segments, and so discards the visual and geometric evidence entirely; DAAAM~\cite{gorlo2026describe} retrieves by matching per-node descriptions, GraphEQA~\cite{saxena2025grapheqa} by traversing a semantic scene graph, and STaR~\cite{yuan2026star} by task-conditioned retrieval over saved keyframes, textual descriptions, and object point clouds. In all cases the information granularity is fixed when the memory is built. 3D-Mem~\cite{yang20253d} and Enter the Mind Palace~\cite{ginting2025enter} retain raw images, are constrained to pre-recorded views, and segment objects in advance, which makes queries about off-list objects unanswerable. Our method runs object segmentation only upon receiving a query, which makes it easy to adjust object granularity without being fixed to predefined lists, utilizing the photorealistic novel view synthesis capabilities of 3DGS.

Beyond memory representations, existing works also diverge in their evaluation benchmarks. EM-EQA benchmarks~\cite{majumdar2024openeqa} predominantly assume single-session explorations of static scenes, overlooking temporal evolution entirely. Other temporal benchmarks, such as NaVQA~\cite{anwar2025remembr}, assess reasoning over continuous navigation videos, focusing on \emph{when} and \emph{where} an event was observed along a continuous trajectory rather than how the environment altered between visits. While recent efforts begin to investigate dynamic environments, Situat3DChange~\cite{liu2026situat3dchange} is confined to change understanding between paired static 3D point scans, whereas Enter the Mind Palace~\cite{ginting2025enter} focuses on multi-episode navigation in simulated settings with limited scene scale and question diversity.

\begin{figure*}[t]
    \vspace*{2mm}%
    \centering
    \includegraphics[width=\textwidth]{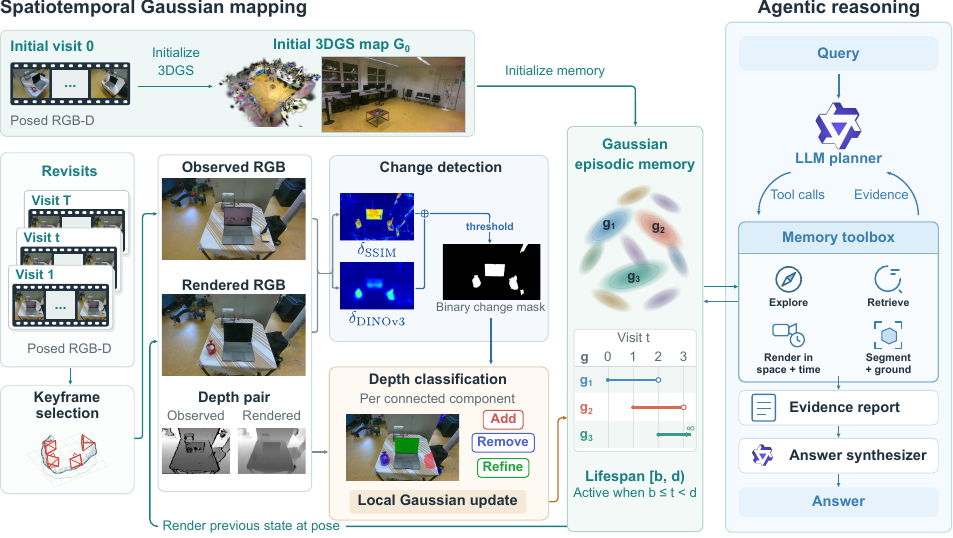}
    \caption{\textbf{Overview of \ours.} \emph{Left:} mapping. On each revisit,
    keyframes rendered from the previous map are compared against the incoming
    observations to obtain a change mask, components of which are labeled as
    \textit{Add}, \textit{Remove}, or \textit{Refine} based on the depth error. The
    affected Gaussian primitives are re-optimized. Each Gaussian carries a lifespan
    $[b,d)$, so the state of scene at any past time can be retrieved as a temporal slice. \emph{Right:}
    reasoning. An LLM planner answers a query through memory tools, and an answer
    synthesizer concludes from the gathered evidence.}
    \label{fig:method_overview}
\end{figure*}

\section{Method}
\label{sec:method}

The goal of this work is to construct a photorealistic spatiotemporal memory of a semi-static scene observed over multiple revisits. The memory supports spatial and temporal reasoning and question answering over the scene's history. As illustrated in \cref{fig:method_overview}, \ours consists of two main components. First, the mapping component initializes a Gaussian splatting map and incrementally maintains its history by detecting and updating changes. This map represents a photorealistic episodic memory capable of on-demand rendering of query-related views and 
object segmentation
across time and space (\cref{subsec:map}). Second, the agentic reasoning component analyzes a query and iteratively gathers grounded evidence from the memory using retrieval, rendering, and segmentation tools
(\cref{subsec:eqa}).
It then synthesizes the final answer from the collected evidence.

\subsection{Spatiotemporal Gaussian Map}
\label{subsec:map}

We consider a robot that observes a semi-static environment over $T+1$ revisits. For each revisit $t$, the input is a posed RGB-D sequence:
\begin{equation}
\mathcal{O}^t =
\left\{
\left(
\mathbf{I}_i^t,
\mathbf{D}_i^t,
\mathbf{K}_i^t,
\mathbf{P}_i^t
\right)
\right\}_{i=1}^{n_t},
\qquad
t \in \{0,\ldots,T\}.
\end{equation}

Where $\mathbf{I}_i^t$ and $\mathbf{D}_i^t$ denote the RGB image and depth map, $\mathbf{K}_i^t$ denotes the camera intrinsics, $\mathbf{P}_i^t \in \mathrm{SE}(3)$ denotes the camera pose in a common world coordinate frame, $n_t$ denotes the number of frames in revisit $t$.

From the first visit $\mathcal{O}^0$, we construct the initial 3D Gaussian map $\mathcal{G}^0$ using gsplat~\cite{ye2025gsplat}. We follow the loss formulation of GaME~\cite{yugay2026gaussian}, including its photometric and geometric terms, but replace its isotropic regularizer with a normalized, thresholded penalty on the per-axis scales,

\begin{equation}
L_{\text{scale}} = \frac{1}{|\Omega| + \epsilon} \sum_{(i,k) \in \Omega} \max \left( 0, \frac{|s_{i,k} - \bar{s}_i|}{\bar{s}_i + \epsilon} - \tau_{\text{scale}} \right),
\end{equation}

which penalizes extreme anisotropic elongation of the Gaussians.

\subsubsection{Cross-Visit Change Detection}
\label{subsubsec:change}

Given the next revisit $\mathcal{O}^t$, we first select a spatially diverse subset of keyframes. A frame $i$
is considered keyframe if its pose distance $d(\mathbf{P}_i^t, \mathbf{P}_j^t)$ from every previously selected keyframe $j$ exceeds $\tau_{\mathrm{pose}}$. We denote the indices of the selected keyframes by $\mathcal{K}^t$. For each $i \in \mathcal{K}^t$, we render a color image $\bar{\mathbf{I}}_i^{t-1}$ and a depth image $\bar{\mathbf{D}}_i^{t-1}$ from the previous Gaussian map $\mathcal{G}^{t-1}$ at the pose of the current observation. 
Because the new observation and the rendering share the same camera pose, their differences primarily capture changes in the scene rather than in the viewpoint.
Then we score both visual and semantic changes. The visual term is the SSIM similarity map between the observed and the rendered images, and the semantic term is the cosine similarity between their DINOv3~\cite{simeoni2025dinov3} feature maps $\mathbf{f}_i^t$ and $\bar{\mathbf{f}}_i^{t-1}$, bilinearly upsampled from patch to pixel resolution. %
Since both terms quantify similarity, the change score $\mathbf{c}_i^t$ is defined as their complement:
\begin{equation}
\label{eq:change_score}
\mathbf{c}_i^t = 1 - [w \ \mathrm{SSIM}\!\left( \bar{\mathbf{I}}_i^{t-1}, \mathbf{I}_i^t \right) + (1-w) \cos\!\left( \bar{\mathbf{f}}_i^{t-1}, \mathbf{f}_i^t \right)],
\end{equation}
with $w = 0.3$. Both similarity terms are clamped to $[0,1]$, so that $\mathbf{c}_i^t \in [0,1]^{H \times W}$. Thresholding it yields the binary change mask $\mathbf{M}_i^t = \mathbb{I}[\mathbf{c}_i^t > \tau_s]$.

Using the observed and the rendered depth, we classify each connected component $\mathcal{S}$ of $\mathbf{M}_i^t$ into three types of general object movement: \textit{Add}, \textit{Remove}, and \textit{Refine}. We compute the relative depth error $\delta \mathbf{D}_i^t = ( \bar{\mathbf{D}}_i^{t-1} - \mathbf{D}_i^t ) / \mathbf{D}_i^t$. A positive error indicates newly observed geometry in front of the preceding map, whereas a negative error indicates that previously observed foreground geometry is no longer present. We therefore compute the fractions
\begin{equation}
\label{eq:fractions}
r_{\mathrm{add}}(\mathcal{S}),\, r_{\mathrm{remove}}(\mathcal{S})
=
\frac{1}{|\mathcal{S}|}
\sum_{\mathcal{S}}
\mathbb{I}\!\left[ \pm\, \delta \mathbf{D}_i^t > \tau_d \right],
\end{equation}
where $\mathbb{I}[\cdot]$ is the indicator function and the positive and negative signs
give $r_{\mathrm{add}}$ and $r_{\mathrm{remove}}$, respectively.

The \textit{Add} operation is triggered if $r_{\mathrm{add}}(\mathcal{S})>\tau_m$ and the \textit{Remove} operation is triggered if $r_{\mathrm{remove}}(\mathcal{S})>\tau_m$. Both operations may be triggered for the same component, as commonly occurs when an object is relocated. If neither condition is satisfied, the component is assigned a \textit{Refine} operation to account for appearance changes or local reconstruction errors, thereby maintaining high-fidelity rendering over a long period of time.

\subsubsection{Local Map Updates and Temporal State Storage}
\label{subsubsec:delta}

We adopt distinct selection and initialization strategies tailored to the three mask types:

\textit{Refine Mask:} We project Gaussian primitives onto the camera plane and set the per-pixel depth threshold to $\tau_{\text{refine}}(\mathbf{u}) = \max\!\left( \bar{\mathbf{D}}_i^{t-1}(\mathbf{u}), \mathbf{D}_i^t(\mathbf{u}) \right)$. Gaussian primitives that fall within the mask and lie closer to the camera than $\tau_{\text{refine}}$ are selected and denoted as $\mathcal{G}^{t-1}_{\text{refine}}$.

\textit{Add Mask:} We back-project the pixels within the mask into 3D space using $\mathbf{D}_i^t$, and follow gsplat~\cite{ye2025gsplat} to initialize the parameters for these new Gaussian primitives. These newly created primitives are added to the set $\mathcal{G}^{t}_{\text{new}}$.

\textit{Remove Mask:} We apply the same selection strategy as in the \textit{Refine Mask}, but set the depth threshold to $\tau_{\text{remove}}(\mathbf{u}) = \mathbf{D}_i^t(\mathbf{u})$. The selected primitives are directly deleted from the map. Subsequently, we sample new Gaussians within these masked regions using the same sampling method to fill potential disocclusion holes caused by the removal of foreground objects. These newly sampled Gaussians are also added to $\mathcal{G}^{t}_{\text{new}}$.

The remaining unselected primitives are designated as static Gaussians, denoted by $\mathcal{G}^{t-1}_{\text{static}}$. To accelerate the optimization procedure, we only update the active primitives in $\mathcal{G}^{t}_{\text{new}} \cup \mathcal{G}^{t-1}_{\text{refine}}$, while keeping $\mathcal{G}^{t-1}_{\text{static}}$ frozen, however, all primitives still participate in rendering. The active Gaussians are optimized using the same loss formulation as during initial map construction (\cref{subsec:map}), but excluding the scale regularization term $L_{\text{scale}}$. Because each keyframe's Gaussians are optimized for only 100 iterations, the horizon is too short for significant shape distortion to accumulate, and enforcing the regularizer over so few steps merely adds computational overhead while prematurely constraining reconstruction quality.

To avoid saving one map $\mathcal{G}^{t}$ per revisit, we keep a single set $\mathcal{G}$ of all Gaussian primitives ever created and attach to each primitive $g$ a lifespan $[b_g, d_g)$, where $b_g$ is the revisit at which $g$ was added and $d_g$ the revisit at which it was removed, with $d_g = \infty$ while $g$ is still present.
An update writes only these two fields: primitives in $\mathcal{G}^{t}_{\text{new}}$ are appended with $b_g = t$, those selected by \textit{Remove} are closed with $d_g = t$, and those in $\mathcal{G}^{t-1}_{\text{refine}}$ are closed with $d_g = t$ while their re-optimized copies are appended with $b_g = t$. The map at any revisit is then recovered by a temporal slice,
\begin{equation}
\label{eq:slice}
\mathcal{G}^t = \left\{ g \in \mathcal{G} \mid b_g \le t < d_g \right\}.
\end{equation}
Thus, the representation grows with accumulated changes rather than repeated visits.
We iterate these steps for every subsequent revisit $t \in \{1,\ldots,T\}$. The resulting spatiotemporal memory retains only the Gaussian primitives $\mathcal{G}$ and the keyframe poses of each revisit:
\begin{equation}
\label{eq:memory}
\mathcal{M} = \left( \mathcal{G}, \;
\left\{ \mathbf{P}_i^t \mid t \in \{0,\ldots,T\},\, i \in \mathcal{K}^t \right\} \right).
\end{equation}

\subsection{Agentic Spatiotemporal Question Answering}
\label{subsec:eqa}

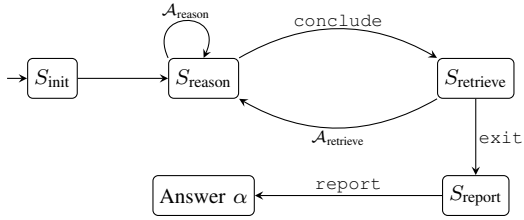
\begin{figure}[t]
\centering
\begin{tikzpicture}[
  >=stealth, font=\footnotesize, node distance=0pt,
  st/.style={draw, rounded corners=2pt, inner sep=2.5pt, minimum height=15pt, align=center},
  lab/.style={font=\fontsize{7}{8}\selectfont\ttfamily, inner sep=1pt},
  lab2/.style={font=\fontsize{6.5}{7}\selectfont, inner sep=1pt}]
  \node[st] (init) at (0,0) {$S_{\text{init}}$};
  \node[st] (reason) at (2.0,0) {$S_{\text{reason}}$};
  \node[st] (retr) at (5.6,0) {$S_{\text{retrieve}}$};
  \node[st] (rep) at (5.6,-1.55) {$S_{\text{report}}$};
  \node[st] (ans) at (2.0,-1.55) {Answer $\alpha$};
  \draw[->] (-0.6,0) -- (init);
  \draw[->] (init) -- (reason);
  \draw[->] (reason.north east) to[bend left=30] node[above,lab]{conclude} (retr.north west);
  \draw[->] (retr.south west) to[bend left=30] node[below,lab2]{$\mathcal{A}_{\text{retrieve}}$} (reason.south east);
  \draw[->] (reason.north west) .. controls +(-0.3,0.65) and +(0.3,0.65) .. node[above,lab2]{$\mathcal{A}_{\text{reason}}$} (reason.north);
  \draw[->] (retr) -- node[right,lab]{exit} (rep);
  \draw[->] (rep) -- node[above,lab]{report} (ans);
\end{tikzpicture}
\caption{Execution state machine. Each state exposes only its tool group ($\mathcal{A}_{\text{reason}}$ or $\mathcal{A}_{\text{retrieve}}$) plus the control tool on each outgoing edge.}
\label{fig:fsm}
\end{figure}

Given a query $q$ and our episodic memory $\mathcal{M}$, we instantiate an agentic reasoning pipeline consisting of an LLM~\cite{qwen35} planner $\pi$ that iteratively calls a set of memory-interaction tools to collect query-related evidence. At each iteration $k$, the planner selects a tool
$a_k$ and its arguments $\mathbf{z}_k$ based on the query and the evidence
collected so far:
\begin{equation}
(a_k,\mathbf{z}_k) = \pi(q,\mathcal{E}_{1:k-1}), \quad
\mathbf{e}_k = \mathcal{T}_{a_k}(\mathcal{M},\mathbf{z}_k),
\end{equation}
where $\mathcal{T}_{a_k}$ is the selected tool, $\mathbf{e}_k$ its returned evidence, and $\mathcal{E}_{1:k-1} = \{\mathbf{e}_1,\ldots,\mathbf{e}_{k-1}\}$ the evidence gathered so far. The available tools are restricted by an execution state machine (\cref{fig:fsm}). The process continues until the planner terminates the search or reaches a predefined tool-call budget.

To speed up the agent's search in episodic memory $\mathcal{M}$, we render an RGB frame for every keyframe in $\mathcal{K} = \bigcup_{t} \mathcal{K}^t$ from $\mathcal{G}$, and cache its OWLv2~\cite{minderer2023scaling} feature. Moreover, we group the keyframe poses of each revisit into spatial regions $\mathcal{C}^t$ by hierarchical agglomerative clustering~\cite{mullner2011modern}. This enables the agent to search for concepts at two levels of granularity in the memory.

Agentic tools interact with $\mathcal{M}$ by rendering query-related color and depth frames, $\mathcal{R}(\mathcal{G}^t,\mathbf{P}) \mapsto (\hat{\mathbf{I}}, \hat{\mathbf{D}})$, from query-selected poses, they fall into two groups:

\emph{Retrieval tools} ($\mathcal{A}_{\text{retrieve}}$) locate relevant views: \texttt{sweep} gives a global scene overview from $\mathcal{C}^t$; \texttt{search} retrieves query-relevant keyframe poses by OWLv2~\cite{minderer2023scaling} similarity to the cached features and renders one candidate per relevant viewpoint cluster; \texttt{covisible} prunes candidates by volumetric covisibility with a target frame; and \texttt{recall} re-renders the cited visual evidence in the report. 

\emph{Reasoning tools} ($\mathcal{A}_{\text{reason}}$) inspect a located region in detail: \texttt{time\_view} renders one camera pose across selected revisits for viewpoint-aligned comparison; \texttt{relook} refines a pose to clarify an ambiguous region; \texttt{locate} segments a queried entity with SAM3~\cite{carion2026sam} and lifts its mask to metric 3D; \texttt{obj\_view} renders further views of a localized entity; \texttt{compare} compares two localized instances; and \texttt{track} associates instances across revisits.

To structure decision-making and prevent early termination, the agentic pipeline is designed as a finite-state machine, in which each state restricts which tools are available (\cref{fig:fsm}). The agent starts in $S_{\text{init}}$ with global retrieval (\texttt{sweep}, \texttt{search}), then alternates between detailed reasoning ($S_{\text{reason}}$) and retrieval ($S_{\text{retrieve}}$); one pass through both is a \emph{retrieval-reasoning round}. \texttt{conclude} ends a round and purges raw visual context from the context history to limit the number of tokens used and speed up the process; \texttt{exit} ends the search; and \texttt{report} compiles the evidence report.

The report, together with all cited visual and 3D evidence $\mathcal{E}$, is passed to the answer synthesizer $\alpha$, which returns the answer $A=\alpha(q,\mathcal{E})$, a rationale, and a confidence score.

\section{Dataset}
\label{sec:dataset}
We present \oursdata{}, a spatiotemporal question-answering benchmark focused on reasoning across discrete revisits to the same semi-static environment. It comprises multi-visit RGB-D trajectories from real-world and synthetic environments, camera poses, object-level change annotations, and question-answer pairs.
The real-world split comprises three indoor environments, each with four visits, captured using an Intel RealSense D455f camera, estimates camera poses using COLMAP~\cite{schonberger2016structure}, and contains a total of 244 question-answer (QA) pairs. The synthetic split comprises 12 HSSD scenes~\cite{khanna2024hssd} simulated using Habitat-Sim~\cite{savva2019habitat}, with at least three visits per scene. It includes 97 annotated object-level changes and 229 QA pairs. Across both splits, all scene changes and questions are manually verified by a human expert to ensure correct ground-truth answers and unambiguous visual references.

We categorize the benchmark questions into five distinct classes: (1)~\textbf{Counting}, covering object quantities and event frequencies; (2)~\textbf{Movement and Presence}, addressing spatial relocation and object appearance, disappearance, or persistence over time; (3)~\textbf{State Change}, referring to non-spatial property changes such as articulation, orientation, color, or configuration; (4)~\textbf{Static}, comprising questions that can be answered from a single visit without temporal context; and (5)~\textbf{Negative}, comprising questions that test robustness to false premises, incorrectly presumed changes, or events that did not occur. Across all 473 questions, the distribution is $30.7\%$ Negative, $22.8\%$ Movement and Presence, $20.5\%$ State Change, $16.9\%$ Static, and $9.1\%$ Counting.

\begin{table}[t]
\centering
\renewcommand{\arraystretch}{0.95}
\caption{Reconstruction quality and update cost of scene-update methods. Time is the wall-clock cost of a single map update.}
\label{tab:recon}
\resizebox{\columnwidth}{!}{%
\begin{tabular}{lccccc}
\toprule
Method & PSNR [dB] $\uparrow$ & SSIM $\uparrow$ & LPIPS $\downarrow$ & Depth L1 [cm] $\downarrow$ & Time [min] $\downarrow$\\
\midrule
\multicolumn{6}{l}{\cellcolor{gray!12}\textbf{(a) Evaluation on GaME~\cite{yugay2026gaussian} Aria Dataset}} \\
O-SCD~\cite{galappaththige2026changes} & \textbf{32.32} & \textbf{0.927} & \textbf{0.121} & -- & \underline{2.79} \\
GaME~\cite{yugay2026gaussian} & \underline{31.12} & \underline{0.916} & \underline{0.122} & \textbf{1.15} & 29.40 \\
\textbf{GEMS (ours)}          & 29.23 & 0.904 & 0.128 & \underline{2.28} & \textbf{1.12} \\
\midrule
\multicolumn{6}{l}{\cellcolor{gray!12}\textbf{(b) Evaluation on CL-Splats Real-World~\cite{ackermann2025cl} w/ DA3~\cite{lin2025depth} Depth}} \\
O-SCD~\cite{galappaththige2026changes} & \textbf{32.04} & \textbf{0.943} & \textbf{0.056} & -- & \underline{2.03} \\
GaME~\cite{yugay2026gaussian} & 24.25 & 0.887 & 0.134 & \underline{10.79} & 29.81 \\
\textbf{GEMS (ours)}                   & \underline{27.87} & \underline{0.901} & \underline{0.089} & \textbf{9.38} & \textbf{1.67} \\
\bottomrule
\end{tabular}%
}
\end{table}

\begin{table*}[ht!]
\vspace{3mm}
    \footnotesize
    \centering
    \renewcommand{\arraystretch}{0.95}
    \caption{3D object grounding results on the Nr3D dataset~\cite{achlioptas2020referit_3d}. 
    Accuracy (\%) by query characteristics (IoU threshold: 10 \%)
    } \vspace{-8pt}
    \resizebox{\textwidth}{!}{
    \begin{tabular}{l ccc ccc ccc}
    \toprule
      \multicolumn{1}{c}{} &
      \multicolumn{1}{c}{Overall} &
      \multicolumn{1}{c}{w Spatial Lang.} &
      \multicolumn{1}{c}{w/o Spatial Lang.} &
      \multicolumn{1}{c}{w Color Lang.} &
      \multicolumn{1}{c}{w/o Color Lang.} &
      \multicolumn{1}{c}{w Shape Lang.} &
      \multicolumn{1}{c}{w/o Shape Lang.} &
      \multicolumn{1}{c}{w Target Mention} &
      \multicolumn{1}{c}{w/o Target Mention} \\
    \textbf{Method} & IoU$_{\ge 0.10}$ & IoU$_{\ge 0.10}$ & IoU$_{\ge 0.10}$ & IoU$_{\ge 0.10}$ & IoU$_{\ge 0.10}$ & IoU$_{\ge 0.10}$ & IoU$_{\ge 0.10}$ & IoU$_{\ge 0.10}$ & IoU$_{\ge 0.10}$ \\
    \midrule
    OpenFusion~\cite{yamazaki2024open}  & 10.7 & 8.9 & 22.3 & 11.8 & 10.5 & 9.8 & 10.9 & 11.3 & 4.9 \\
    ConceptGraphs~\cite{gu2024conceptgraphs} & 16.0 & 15.0 & 22.3 & 17.6 & 15.7 & 10.8 & 16.9 & 16.9 & 6.6 \\
    BBQ~\cite{linok2025beyondbarequeries}  & 28.3 & 28.1 & 29.8 & 25.2 & 29.0 & 34.3 & 27.3 & 29.6 & 14.8 \\
    KeySG~\cite{werby2025keysg}  & \underline{49.8} & \underline{49.2} & \underline{57.0} & \underline{48.3} & \underline{50.7} & \underline{50.0} & \underline{50.3} & \underline{52.6} & \textbf{25.4} \\
    \textbf{\ours (ours)}  & \textbf{58.9} & \textbf{57.0} & \textbf{71.1} & \textbf{63.2} & \textbf{57.9} & \textbf{58.7} & \textbf{58.9} & \textbf{62.2} & \underline{23.4} \\
  \bottomrule
  \end{tabular}
    }
    \label{tab:obj_ret_com}
\end{table*}

\begin{figure*}[t]
\vspace{1mm}
    \centering
    \includegraphics[width=\textwidth]{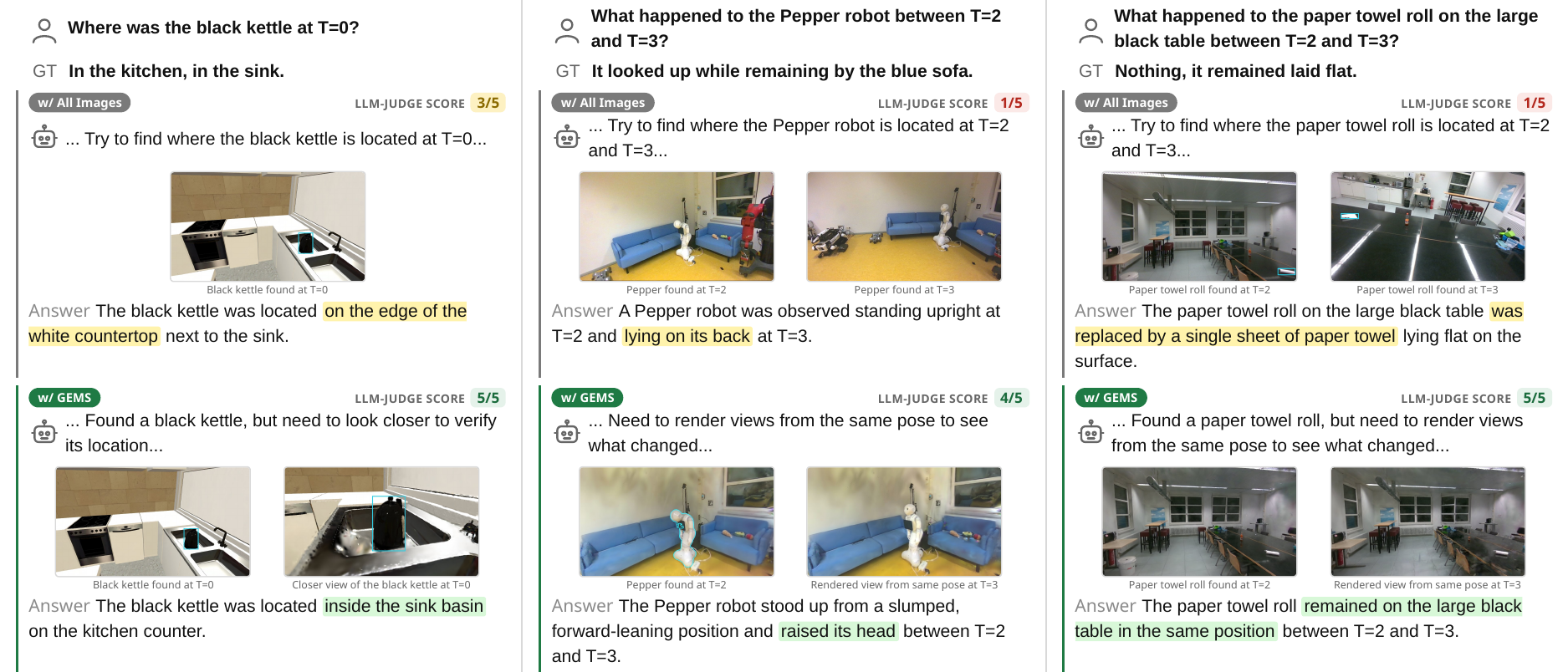}
    \caption{\textbf{Qualitative comparison of \ours against \ours-all-images.} Each column shows a question, the evidence each agent collected, and the LLM-judge score. \emph{Left:} a synthesized closer view resolves a spatial ambiguity from an oblique camera angle. \emph{Middle and right:} \texttt{time\_view} lets \ours compare both visits from one pose, where the baseline is restricted to two different stored viewpoints.}
    \label{fig:qualitative}
\end{figure*}

\section{Experiments}
\label{sec:exp}

\subsection{Reconstruction Quality}
\label{subsec:recon}

We assess whether \ours preserves rendering quality while updating the map incrementally, holding out every 10th input image as the test set, following the same evaluation protocol as GaME~\cite{yugay2026gaussian}.
We compare against two dedicated scene-update methods: GaME~\cite{yugay2026gaussian}, which uses depth to prune obsolete Gaussians and initialize primitives for newly introduced objects, and O-SCD~\cite{galappaththige2026changes}, which updates the map from RGB-only change detection and therefore neither consumes nor predicts depth. We evaluate on two real-world benchmarks with one temporal update per scene: GaME-Aria~\cite{yugay2026gaussian}, with two indoor scenes and dense posed RGB-D sequences, and CL-Splats Real-World~\cite{ackermann2025cl}, with five scenes, sparser views, and no ground-truth depth. Since \ours and GaME both require depth, we synthesize depth for CL-Splats using Depth Anything 3~\cite{lin2025depth}. We report PSNR, SSIM~\cite{wang2004image}, and LPIPS~\cite{zhang2018unreasonable} for photometric quality, depth L1 error for geometric accuracy, and the wall-clock time of a single map update.

\cref{tab:recon} shows that \ours preserves rendering quality on par with other methods while updating the map far more cheaply: an update takes about one minute on both benchmarks, between $18\times$ and $26\times$ faster than GaME. \ours achieves the lowest depth error on CL-Splats and the second lowest on GaME-Aria, and stays within a few dB PSNR of the best photometric scores, which O-SCD achieves without producing any depth at all.

\subsection{Object Grounding from Natural-Language Queries}
\label{subsec:nr3d}

To evaluate \ours on grounding objects from natural-language queries in an indoor scene, we use the Nr3D dataset~\cite{achlioptas2020referit_3d}, which contains single-visit indoor scenes paired with a diverse set of natural-language queries grouped by query type. We adopt the same evaluation protocol and scene identifiers as BBQ~\cite{linok2025beyondbarequeries}. \cref{tab:obj_ret_com} compares \ours against recent methods that build structured representations, reporting grounding accuracy at $\mathrm{IoU} \ge 0.1$.

\ours achieves the best overall accuracy of $58.9\%$ and outperforms all baselines in every query category except queries without a target mention. Both \ours and KeySG~\cite{werby2025keysg} are substantially more accurate than the remaining methods, since both exploit (raw or rendered) visual evidence rather than abstracted object nodes. While KeySG retains a slight edge on queries that omit target mention, \ours achieves superior performance across all other spatial, color, and shape categories.

\begin{table}[t]
\vspace{2mm}
\centering
\renewcommand{\arraystretch}{0.95}
\caption{Blocks (a) and (b): spatiotemporal EQA accuracy from an LLM judge (0--100\%) by question category. Block (c): NVS tool ablation on the questions where the full agent invoked NVS.}
\label{tab:stqa}
\resizebox{\columnwidth}{!}{%
\begin{tabular}{lccccc|c}
\toprule
Method & Count. & Move./Pres. & Neg. & State Ch. & Static & Overall \\
\midrule
\multicolumn{7}{l}{\cellcolor{gray!12}\textbf{(a) Real-world benchmark}} \\
Blind-Qwen3.5~\cite{anwar2025remembr} & 26.7 & 21.2 & 28.3 & 18.7 & 33.1 & 25.0 \\
ReMEmbR~\cite{anwar2025remembr} & 26.7 & 25.5 & 38.6 & 26.2 & 17.6 & 28.3 \\
DAAAM~\cite{gorlo2026describe} & 10.8 & 22.8 & 40.1 & 25.4 & 24.3 & 27.1 \\
Mind Palace~\cite{ginting2025enter} & 34.2 & 28.8 & 30.2 & 24.6 & 45.3 & 31.3 \\
STaR~\cite{yuan2026star} & 39.2 & 46.2 & 45.2 & 42.9 & 50.0 & 44.8 \\
\ours-all-images                       & \underline{63.3} & \textbf{59.2} & \textbf{65.4} & \textbf{63.9} & \underline{62.8} & \textbf{63.2} \\

\textbf{\ours (ours)}                  & \textbf{65.0} & \underline{53.3} & \underline{60.3} & \underline{60.7} & \textbf{68.2} & \underline{60.9} \\
\midrule
\multicolumn{7}{l}{\cellcolor{gray!12}\textbf{(b) Synthetic benchmark}} \\
Blind-Qwen3.5~\cite{anwar2025remembr} & 38.5 & 39.1 & 35.4 & 16.9 & 24.4 & 31.8 \\
ReMEmbR~\cite{anwar2025remembr} & 11.5 & 42.3 & 35.4 & 30.9 & 35.5 & 35.3 \\
DAAAM~\cite{gorlo2026describe} & 23.1 & 37.1 & 48.7 & 22.8 & 36.6 & 38.0 \\
Mind Palace~\cite{ginting2025enter} & 15.4 & 25.0 & 31.8 & 32.4 & 58.7 & 34.2 \\
STaR~\cite{yuan2026star} & 23.1 & 54.4 & 26.0 & 37.5 & 27.3 & 35.5 \\
\ours-all-images                       & \underline{48.1} & \underline{64.9} & \underline{65.6} & \textbf{59.6} & \underline{66.9} & \underline{63.8} \\
\textbf{\ours (ours)}                  & \textbf{63.5} & \textbf{71.4} & \textbf{74.7} & \underline{58.1} & \textbf{76.2} & \textbf{71.0} \\
\midrule
\multicolumn{7}{l}{\cellcolor{gray!12}\textbf{(c) NVS tool ablation on questions invoking NVS ($n=245$)}} \\
number of questions                    & 17   & 53   & 87   & 65   & 23   & 245  \\
\ours-no-NVS                           & 52.9 & 59.9 & 65.2 & 53.1 & 48.9 & 58.5 \\
\textbf{\ours (ours)}                  & \textbf{66.2} & \textbf{74.5} & \textbf{65.5} & \textbf{56.9} & \textbf{64.1} & \textbf{65.1} \\
\bottomrule
\end{tabular}%
}
\end{table}

\subsection{Spatiotemporal Question Answering}
\label{subsec:tqa_results}

We evaluate \ours on spatiotemporal question answering using our \oursdata benchmark. For each scene, we first build the episodic memory $\mathcal{M}$, then run the agentic pipeline to explore it and answer every question in that scene. Following OpenEQA~\cite{majumdar2024openeqa}, we score each predicted answer with an LLM matcher on a scale from 1 to 5, where 1 is completely wrong and 5 is a perfect answer, and report the mean score rescaled to $[0,1]$. We use GPT-5.6-Luna as the judge.

We compare against a range of memory representations. DAAAM~\cite{gorlo2026describe} and Enter the Mind Palace~\cite{ginting2025enter} build 3D scene graphs, ReMEmbR~\cite{anwar2025remembr} stores textual descriptions alone, and STaR~\cite{yuan2026star} combines saved keyframes with textual descriptions and object point clouds. We additionally compare to \ours-all-images, a variant that retains all posed RGB-D frames across revisits as its memory and runs the same agentic pipeline, except that it cannot use the NVS tools; it acts as an upper bound on the visual information available to the agent. We also compare against a blind LLM, Qwen3.5~\cite{qwen35}, where the model is purely guessing at the answers to the questions; it serves as a lower bound. For fairness, every method, including \ours, uses Qwen3.5~\cite{qwen35} as its underlying LLM.
\cref{tab:stqa} breaks the results down by query type. \ours attains the best overall score on both splits, $60.9\%$ on the real-world and $71.0\%$ on the synthetic split, against $44.8\%$ and $35.5\%$ for the strongest baseline, STaR. On each question class, \ours outperforms all the other baselines with around $10\%$.

On the real-world split, \ours slightly underperforms \ours-all-images, yet it achieves higher performance on the synthetic split despite storing no images. The per-class breakdown and overall metrics reveal that while \ours-all-images exhibits consistent performance across both splits, \ours shows a noticeable performance gap. Specifically, \ours achieves nearly a $10\%$ improvement on the synthetic split for movement and presence, negative, and static questions, while yielding nearly identical results on both splits for state change and counting questions. On the real-world split, sensor noise and pose estimation errors result in imperfect Gaussian maps. Although the synthesized views remain high quality when close to the training trajectories (\cref{subsec:recon}), our novel view synthesis (NVS) tools often query views far from the training poses, occasionally introducing visual artifacts. These degraded images lower the performance of \ours on the real-world split. Conversely, perfect depth maps and poses in the synthetic split eliminate these artifacts, yielding significant gains. Overall, the performance benefits provided by NVS tools outweigh their potential drawbacks. A detailed isolation study on the impact of NVS is presented in (\cref{subsec:ablation}).

\subsection{Storage Efficiency}
\label{subsec:storage}

\begin{figure}[t]
    \vspace{1mm}
    \centering
    \includegraphics[width=\linewidth]{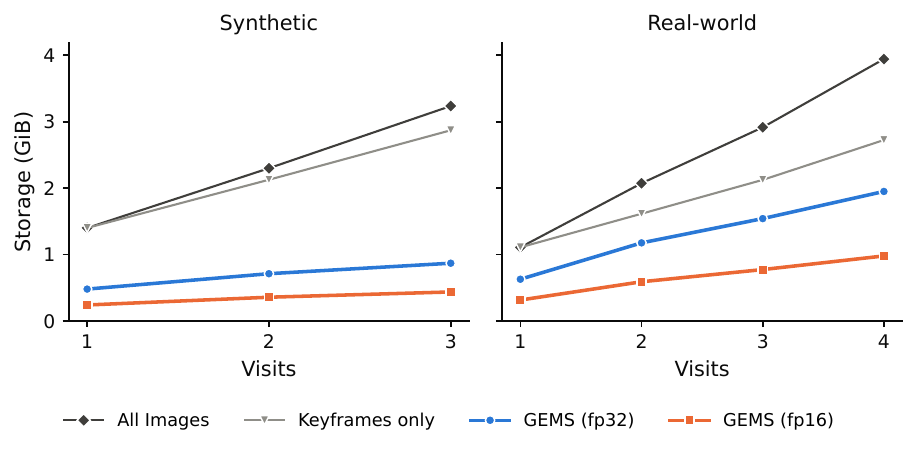}
    \caption{\textbf{Storage growth.} Memory footprint against the number of
    revisits, for one synthetic and one real-world scene of \oursdata. Retaining
    all posed RGB-D frames, or only the keyframes, costs a fixed amount per
    revisit, whereas \ours stores only the Gaussian primitives affected by a
    change. We report \ours at single (fp32) and half (fp16) precision.}
    \label{fig:storage}
\end{figure}

Finally, we compare the growth of \ours's episodic memory against two alternatives that retain images directly: all posed RGB-D frames, and the posed keyframes only. \cref{fig:storage} reports all of them on one synthetic and one real-world scene of \oursdata.
Since \ours re-optimizes and stores Gaussian primitives only where a change is detected, its memory grows in proportion to the change observed at each revisit rather than paying a fixed cost per visit.
After four revisits of the real-world scene, \ours occupies $1.95$~GiB at single precision and $0.98$~GiB at half precision, compared with $2.73$~GiB for the keyframes alone and $3.94$~GiB for the full-frame sequence. The absolute gap widens with each revisit, so \ours is more efficient whenever changes are localized, the common case in semi-static environments.

\subsection{Ablation Studies}
\label{subsec:ablation}

To assess the contribution of NVS, we select the $245$ questions for which the full agent invoked \texttt{time\_view} or \texttt{relook}, and rerun the complete pipeline on them with both tools disabled, using the same Gaussian memory and inference budget. Restricting to these questions separates the agent’s decision to invoke the tools from the tools’ own contribution. Access to the two tools improves the mean QA score by $6.6$ percentage points, with gains across all five categories (\cref{tab:stqa}). This is the benefit of NVS, where the agent elects to use it, not across the whole benchmark. \cref{fig:qualitative} illustrates how viewpoint-aligned rendering supports temporal comparison: restricted to recorded viewpoints, the agent may fail to re-identify an object across perspective changes (right) or mistake a viewpoint difference for a physical state change (middle). Because both tools are removed jointly, the ablation does not separate spatial view refinement from temporal alignment.
\section{Conclusion}
\label{sec:conclusion}

We presented \ours, a persistent spatiotemporal memory built on 3D Gaussian Splatting that stores environmental change as compact temporal deltas and answers episodic-memory questions by rendering novel views across space and time. By updating and storing only the Gaussians in changed areas, \ours achieves superior storage efficiency and is still able to re-render the required images on demand from arbitrary camera poses at any time. Experiments on our long-term spatiotemporal benchmark show that combining spatial viewpoint selection with cross-temporal rendering improves question-answering accuracy and consistently outperforms established baselines.

\section{Acknowledgements}

\footnotesize{
This work was supported by the German Federal Ministry of Research, Technology, and Space (BMFTR) through the Robotics Institute Germany (RIG), and in part by the Deutsche Forschungsgemeinschaft (DFG, German Research Foundation) under Germany's Excellence Strategy -- EXC 2120/2 -- 390831618 --  Cluster of Excellence Integrative Computational Design and Construction for Architecture (IntCDC).
}

\begin{footnotesize}
    \bibliographystyle{IEEEtran}
    \bibliography{references.bib}
\end{footnotesize}

\end{document}